\pdfoutput=1
\documentclass[letterpaper, 10 pt, conference]{ieeeconf}  %

\IEEEoverridecommandlockouts                              %

\usepackage{graphics} %
\usepackage{amsmath} %
\usepackage{amssymb}  %

\usepackage{cite}
\usepackage{array}
\usepackage{caption}
\usepackage{subcaption}
\usepackage{tikz}
    \usetikzlibrary{shapes}
\usepackage{pgfplots}
    \usepgfplotslibrary{colorbrewer}
    \pgfplotsset{compat=newest}
\usepackage{filecontents}
\usepackage[export]{adjustbox}
\usepackage{balance}
\usepackage{breakurl}
\usepackage[breaklinks]{hyperref}

\newcommand{\xxnote}[3]{}
\ifx\hidenotes\undefined
\usepackage{color}
\renewcommand{\xxnote}[3]{\color{#2}{#1: #3}}
\fi

\title{\LARGE \bf
Desk Organization: Effect of Multimodal Inputs on Spatial Relational Learning
}

\author{Ryan Rowe*, Shivam Singhal*, Daqing Yi, Tapomayukh Bhattacharjee, and Siddhartha S. Srinivasa%
\thanks{*These authors contributed equally to the work. 
All the authors are with Paul G. Allen School of Computer Science and Engineering, University of Washington, Seattle, Washington 98195
{\tt\footnotesize \{rfrowe, shivam42, dqyi, tapo, siddh\}@cs.washington.edu}}%
\thanks{This work was funded by the National Institute of Health R01 (\#R01EB019335), National Science Foundation CPS (\#1544797), National Science Foundation NRI (\#1637748), the Office of Naval Research, the RCTA, Amazon, Honda, and the UW Allen School Postdoc Research Award. We thank Aditya Mandalika for help with robot experiments.}}
\renewcommand{\baselinestretch}{0.97}

\begin{document}

\maketitle
\thispagestyle{empty}
\pagestyle{empty}

\begin{abstract}

For robots to operate in a three dimensional world and interact with humans, learning spatial relationships among objects in the surrounding is necessary. Reasoning about the state of the world requires inputs from many different sensory modalities including vision (\textsc{v}) and haptics (\textsc{h}). We examine the problem of desk organization: learning how humans spatially position different objects on a planar surface according to organizational ``preference''. We model this problem by examining how humans position objects given multiple features received from vision and haptic modalities. However, organizational habits vary greatly between people both in structure and adherence. To deal with user organizational preferences, we add an additional modality, ``utility'' (\textsc{u}), which informs on a particular human's perceived usefulness of a given object. Models were trained as generalized (over many different people) or tailored (per person). We use two types of models: random forests, which focus on precise multi-task classification, and Markov logic networks, which provide an easily interpretable insight into organizational habits. The models were applied to both synthetic data, which proved to be learnable when using fixed organizational constraints, and human-study data, on which the random forest achieved over 90\% accuracy. Over all combinations of \{\textsc{h}, \textsc{u}, \textsc{v}\} modalities, \textsc{uv} and \textsc{huv} were the most informative for organization. In a follow-up study, we gauged participants preference of desk organizations by a generalized random forest organization vs. by a random model. On average, participants rated the random forest models as 4.15 on a 5-point Likert scale compared to 1.84 for the random model.
\end{abstract}

\section{Introduction}
\label{sec:introduction}

Researchers have developed robotic systems that can perform a variety of household tasks ranging from automated cleaning robots \cite{fiorini2000cleaning}, to aiding in kitchen tasks \cite{dillmann2004teaching,beetz2008assistive}, to robots assisting the disabled and elderly in their everyday lives \cite{choi2009list}.
The knowledge of encoding spatial relations of objects is relied on by many in-home tasks such as organizing a desk or a shelf, cleaning an area, or retrieving requested objects.

\begin{figure}[ht]
	\centering
	\includegraphics[width=0.49\linewidth]{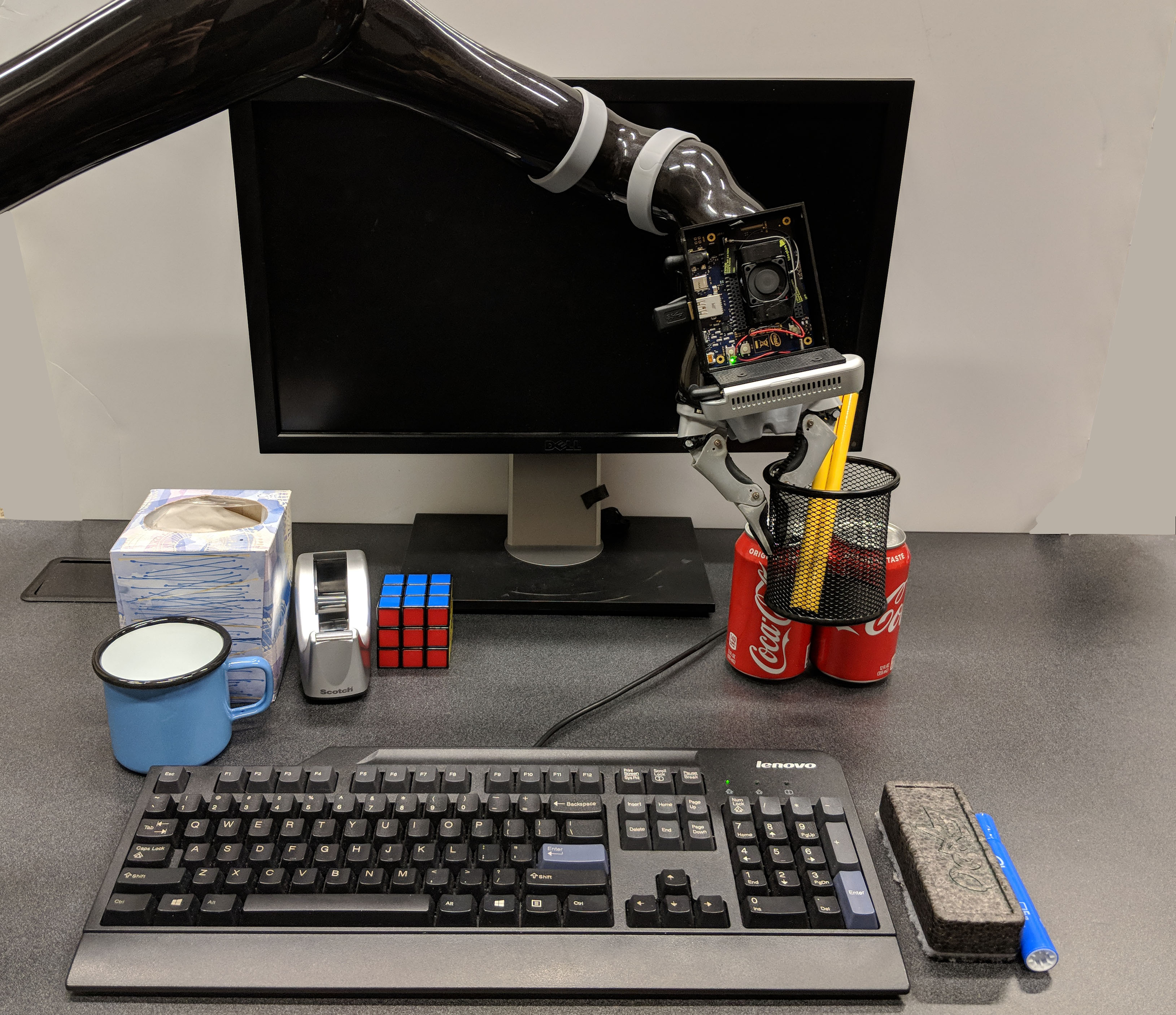}
	\includegraphics[width=0.49\linewidth]{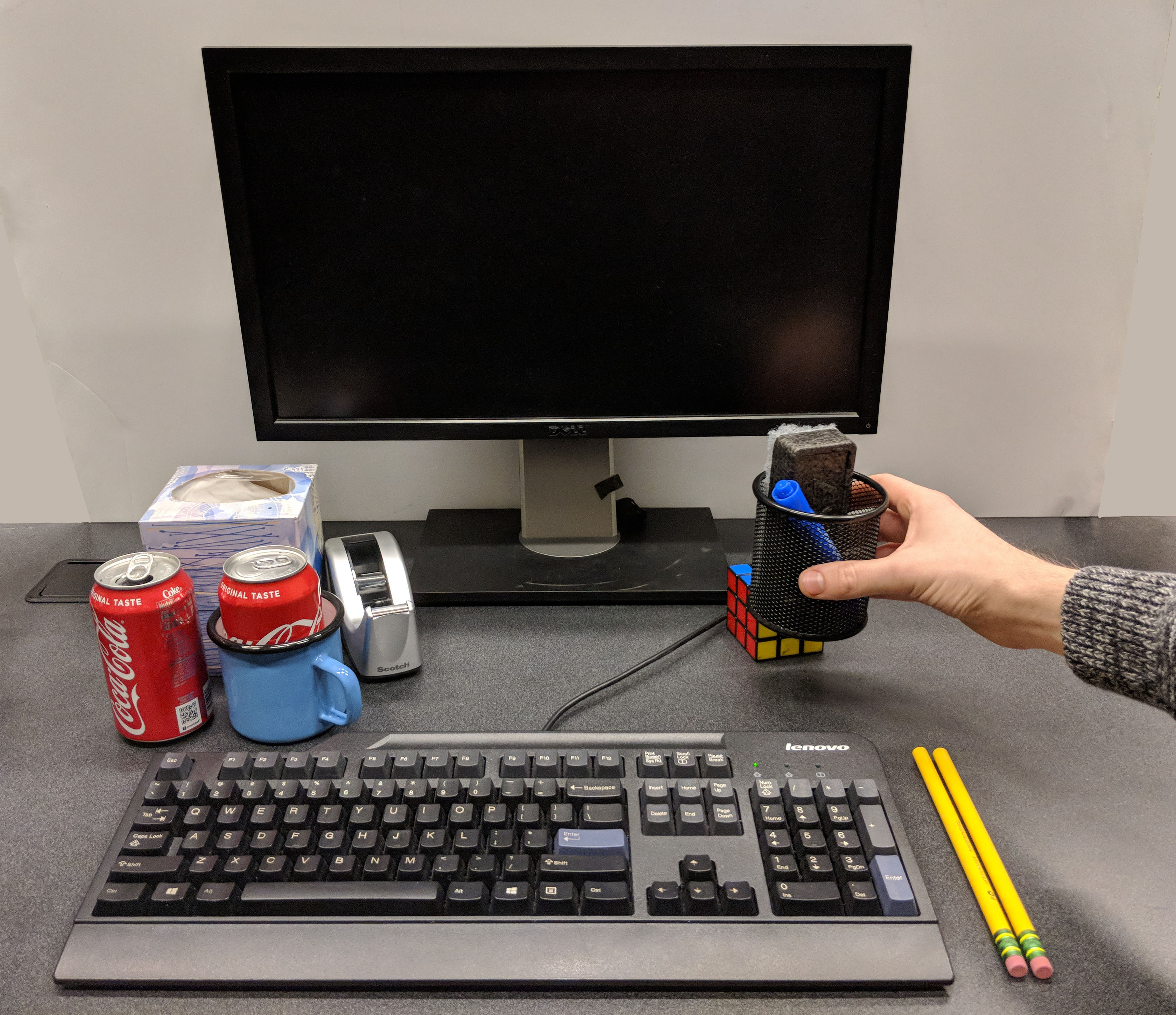}
	\caption{A sample desk organization task.}
	\label{fig:desk_organization}
	\vspace{-0.5cm}
\end{figure}

\begin{figure*}[ht]
    \centering
    \includegraphics[width=\linewidth]{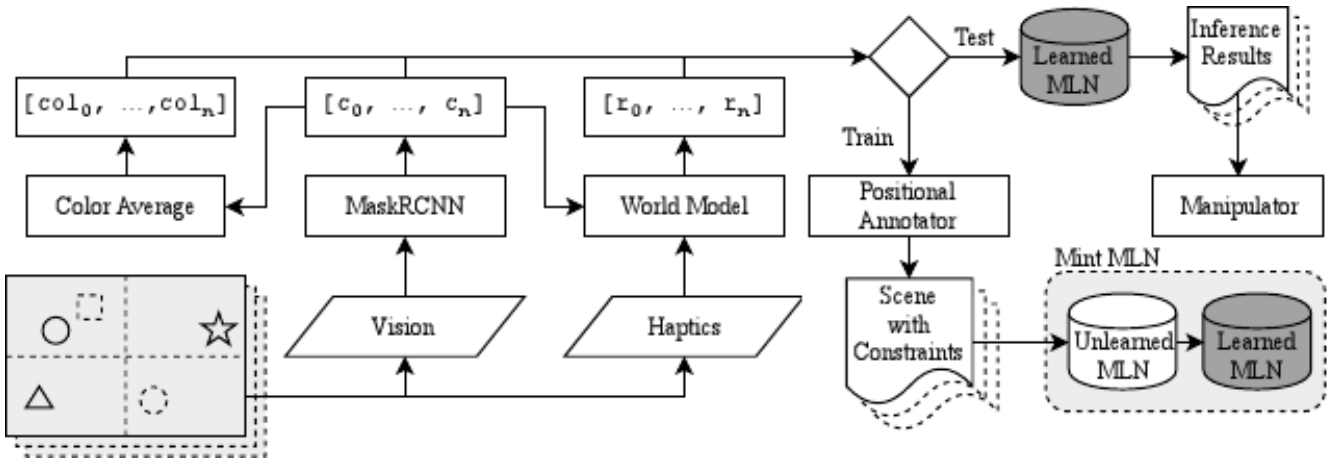}
    \caption{Example pipeline for the desk organization problem. Utilizes MaskRCNN for object detection on visual input which is used to extract clusters ($c_i$). From this visual input, properties such as color and shape can be extracted by annotation models. Using the detected clusters, haptic properties can be gathered through manipulation with a robotic end-effector equipped with haptic sensors.}
    \label{fig:pipeline}
    \vspace{-0.5cm}
\end{figure*}

Spatially organizing objects in turn requires an algorithm to model a person's preferences as well as an object's attributes in order to infer spatial relationships suited to a particular task. In this paper, we investigate this problem further by focusing on the task of autonomous desk organization. Through this task, we are particularly interested in exploring the role of an object's multimodal physical attributes and a person's organizational preferences in learning and inferring spatial relationships.

Desk organization presents a challenge as the location of objects on an organized desk is a function of not only the object's physical attributes but also a person's preferences of what is ``organized'' and also their perceived usefulness of each object.

Multimodal learning, which leverages information from inputs from multiple modalities such as vision and haptics, can provide interesting insights into the roles these inputs play in the task of spatially organizing objects. In this paper, we consider physical attributes such as color, shape, size, weight, and rigidity of an object as well as its functional attributes such as its utility in the context of a user's preference in order to infer spatial relations.

We approached learning these spatial relationships from multimodal inputs using both Markov Logic Networks (MLN)~\cite{richardson2006markov} and Random Forests \cite{Breiman2001}. For visual modality, we trained a MaskRCNN~\cite{he2017mask} over the objects used during experimentation as a proof of concept to detect objects as well as to identify their color and shape. For haptic modality, we used a haptic sensor to obtain values for rigidity. Some physical attributes such as weight (heavy or light), size (large or small), and functional attributes (utility) are subjective and depend on the user. Therefore, we performed human-subject studies and obtained values for these attributes from human responses. Our models use these attributes to learn spatial relationships that a particular user prefers when organizing their desk, then predicts object placements and relationships for new desk situations.

Our results indicate that people do follow latent patterns when organizing objects on a desk. We selected MLN as it provides easily interpretable results in the form of weighted formula, which showed that utility was the most informative modality when determining spatial relationships between objects, followed by vision and lastly haptics. We compared Markov logic networks to a random forest which trades interpretability for increased accuracy, lower training time, and better abstraction of organizational concepts.

Our contributions include:
\begin{enumerate}
	\item Analyzed the role of multimodal inputs in spatial relational learning
	\item Modeled user's preferences in the context of spatial relational learning using a desk organization task
	\item Added utility modality to model individual differences in object preferences
\end{enumerate}

\section{Related Work}
\label{sec:related_work}

\subsection{Multimodal Spatial Reasoning}
Spatial reasoning is a fundamental skill that supports robotic understanding of human intent~\cite{turaga2008machine} and execution of daily tasks~\cite{chung2008daily}. There has been extensive work done in understanding spatial reasoning for robotics \cite{Mansouri2013ARF, spatialReasoningReview, Kennedy2007SpatialRA}. A typical organizational directive, such as ``place the eraser to the left of the keyboard'' not only relies on the robot's ability to properly categorize objects (``eraser'' and ``keyboard'') but also on its learned spatial knowledge associated with prepositions (``left of'')~\cite{paul2016efficient}. Observing how humans define spatial relations in tasks allows spatial relation learning from humans, which has now attracted attention in the fields of computer vision and audition~\cite{webscalengrams,yao2010i2t, bottomuptopdown, areasattentionic} as well as robotics~\cite{thomason2016learning,taniguchi2016221,isobe2017learning}. Maintaining spatial concepts between objects and places~\cite{isobe2017learning} as a form of knowledge enhances the robustness in navigating a human-living environment~\cite{taniguchi2016221} by self-localization of the objects in the domain (e.g. water bottle, box, mouse, etc.). 

Our focus is to analyze how multimodal inputs specifically facilitate spatial reasoning for robotics. When a robot physically interacts with an environment, there is an opportunity to collect data from multiple modalities such as vision, haptics, textures, gestures, language, and audio~\cite{thomason2016learning,turk2014multimodal}. The mutual enhancement between multiple modalities inspires our multimodal learning~\cite{ldamultimodal}. Multimodal learning in robotics does not only provide more information to spatial learning~\cite{sisbot2007spatial}, but also explores the association between different perceptions~\cite{thomason2016learning}. Spatial relational learning falls under the multimodal task of translation, namely, using vision, haptic, and/or audio data to ground the natural language which describes the spatial relations.

\subsection{Relational Models}
Markov Logic Networks (MLN), which are probabilistic graphical models, are a common method to represent relational information about the objects in the world \cite{richardson2006markov}, in this case, spatial relations (e.g., the coffee cup is behind the monitor) and object attribute relations (e.g., the water bottle is a hard blue cylinder). In addition to MLNs, others have tried neural networks, add-or graphs, and support vector machines to varying degrees of success \cite{multimodalmachinelearning, webscalengrams, thomason2016learning}. There are some similar methods which use attention augmented networks to learn spatial relations \cite{ban, attentionSpatial}. MLNs combine the ability of a Bayesian network to encode arbitrary probability distributions with the power of first-order logic \cite{richardson2006markov}. We use the MLN, as such models require less data to train, allow for the addition of hand-crafted features (e.g., for spatial relations), and offer more interpretability than others \cite{multimodalmachinelearning}. Similar to \cite{skovcaj2016integrated}, we use the MLN to encode cross-modal relationships essentially serving as the fusion step in multimodal learning. First order formulae in our MLNs are used to encode spatial-modality relationships between object attributes and spatial relationships while the graph structure allows for cross-modal interaction. In our case, MLN groundings are attribute predicates and domains (such as color, shape, etc.) along with an object and attribute value (blue, rectangle, etc.). Each predicate can, as in \cite{nyga2014pr2}, have an associated annotator which generates groundings in the associated domain. These annotators are separate, independent models which are given detected clusters and produce annotations in their expert domain, such as color, shape, etc. Figure~\ref{fig:pipeline} shows how a MaskRCNN can be used in conjunction with visual annotators, which operate on image input, as well haptic annotators which require visual input to detect and classify rigidity and weight by robotic manipulation.

We also use an ensemble learning algorithm, namely random forest, to compare with the MLN \cite{Breiman2001}. We vectorized the attribute predicates and domains. We were motivated to try these as an alternative to the more interpretable MLNs. By turning spatial reasoning into a classification problem with discrete spatial relations, random forests can be used for spatial reasoning. They have been used in various fields for multi-modal learning. Some examples include classification of Alzheimer's disease \cite{rf_alz}, automatic job-candidate screening based on video CV's \cite{cvpr_rf}, and news article classification \cite{news_rf}. The features used depend on the problem at hand: \cite{rf_alz} uses, amongst others, MRI volumes and voxel-based FDG-PET signal intensities. On the other hand, the job-candidate used videos \cite{cvpr_rf}, while the news article one used n-gram textual features and a representative image \cite{news_rf}. Even so in the case of missing/incomplete data. So, random forests seem to be a pragmatic model to use for our problem.

\section{A Tale of Two Models for Desk Organization}

\begin{figure}
	\centering
	\includegraphics[width=.9\linewidth]{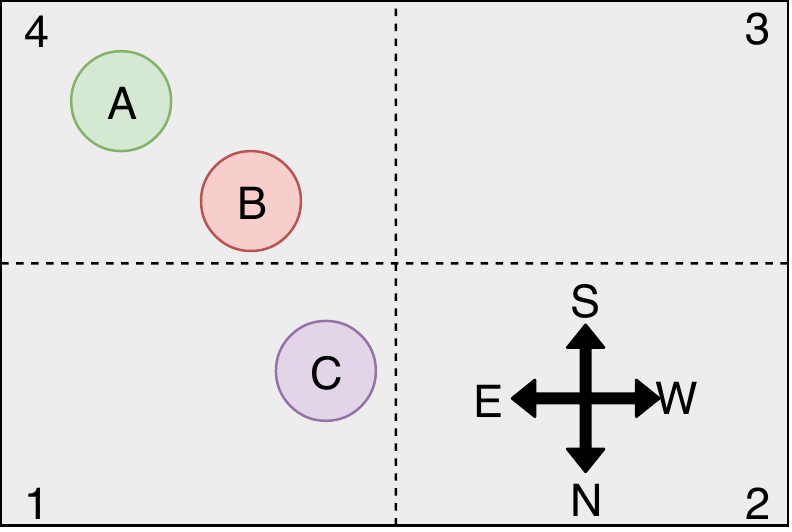}
	\caption{An example scene with 3 objects, A, B, and C, showing their positions relative to a desk (which quadrant the object is in) and relative to other objects (by cardinal direction).}
	\label{fig:spatial_information}
	\vspace{-0.7cm}
\end{figure}

We define the desk organization problem as spatial inference according to the properties of objects~\cite{malone1983people,pantofaru2012exploring}.
For a given desk organization ``scene'', we assume that there are $ K $ objects $ \mathbb{O} = \{ o_1, \cdots , o_K \} $. We also assume that we can extract features  $ F(o_i) $  of an object $ o_i $ using multiple modalities. These features are used to infer the spatial information $ \mathbb{R} $ about how the object should be organized with other objects. We model the human preference as  $ P( \mathbb{R} \mid F(o_1) , \cdots , F(o_K) ) $, which then defines the desk organization problem as 
\begin{equation}
\label{eq:spatial_infer}
\arg\max_{ R } P( \mathbb{R} \mid F(o_1) , \cdots , F(o_K) ).
\end{equation}
so that a robot can organize the objects based on inferred spatial information.

In this paper, we focus on two forms of spatial relations. Objects can be located relative to the desk and relative to each other.
Spatial relations between objects are encoded as cardinal directions between unique object pairs. For example, in Figure~\ref{fig:spatial_information}, object A is southeast of object B (and correspondingly object B is northwest of object A).
Spatial relations between an object and the desk are described by the quadrant $\in \{1,2,3,4\}$ in which the object resides.
In Figure~\ref{fig:spatial_information}, object A is in quadrant $ 4 $.
This quadrant-based spatial relation serves two purposes.
First, it models how people position objects on desks as very few (if any) are placed in the center of a desk (where the quadrants would intersect) and most objects are placed in the corners of the desk.
Second, it simplifies the relational space that the model must learn as the relative relations are known between any two objects in different quadrants (e.g., any object in quadrant 1 is necessarily north of another in quadrant $ 4 $).
In addition to cardinal directions between objects, we allow for smaller objects to be placed ``in'' larger objects. Thus, we define two types of spatial information for the desk organization problem, which are 
\begin{itemize}
	\item $ R^{Quad} (o_i) \in \{ 1, \cdots , 4 \} $ tells which quadrant object $ o_i $ is in.
	\item $ R^{Rel} (o_i, o_j ) \in \{ {\sc E} , {\sc NE}, {\sc N}, {\sc NW}, {\sc W}, {\sc SW}, {\sc S}, {\sc SE}, {\sc IN},$\\${\sc NONE}\} $ tells the spatial relation of object $ o_i $ relative to $ o_j $.
\end{itemize}

\begin{figure}
	\centering
	\begin{subfigure}[t]{0.48\linewidth}
		\centering
	    \includegraphics[width=0.95\textwidth]{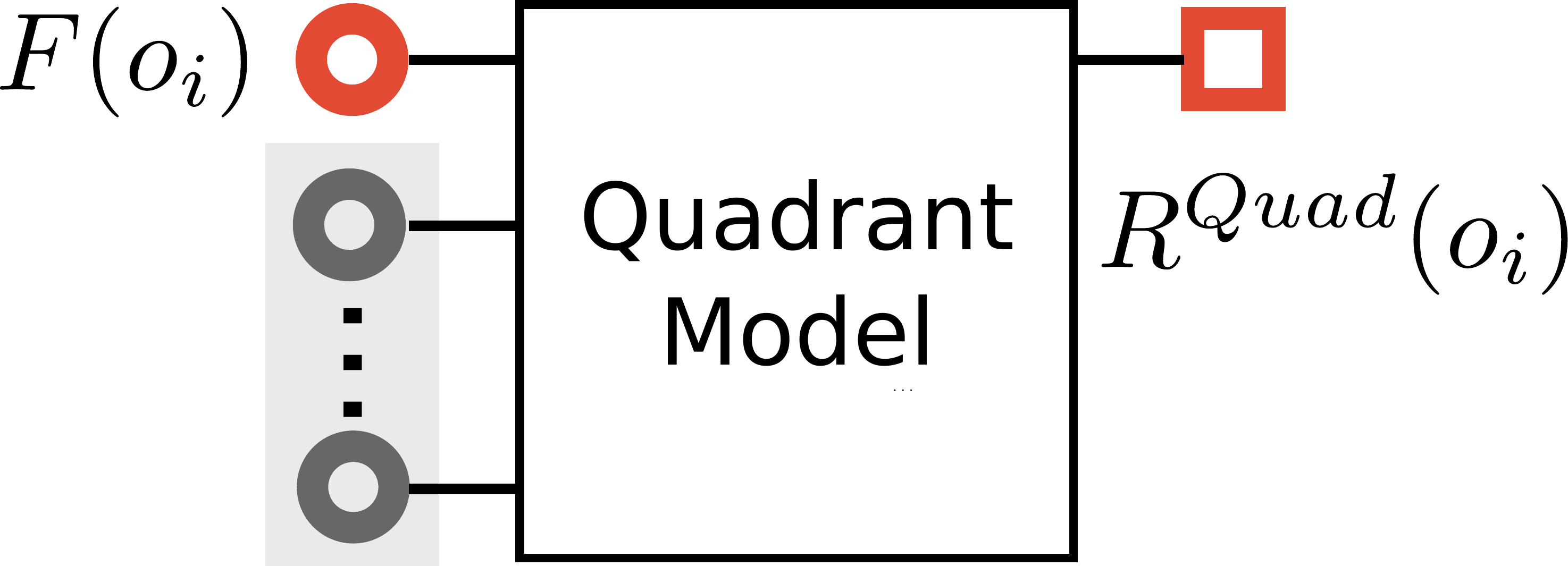}
	\end{subfigure}
	\begin{subfigure}[t]{0.48\linewidth}
		\centering
		\includegraphics[width=0.95\textwidth]{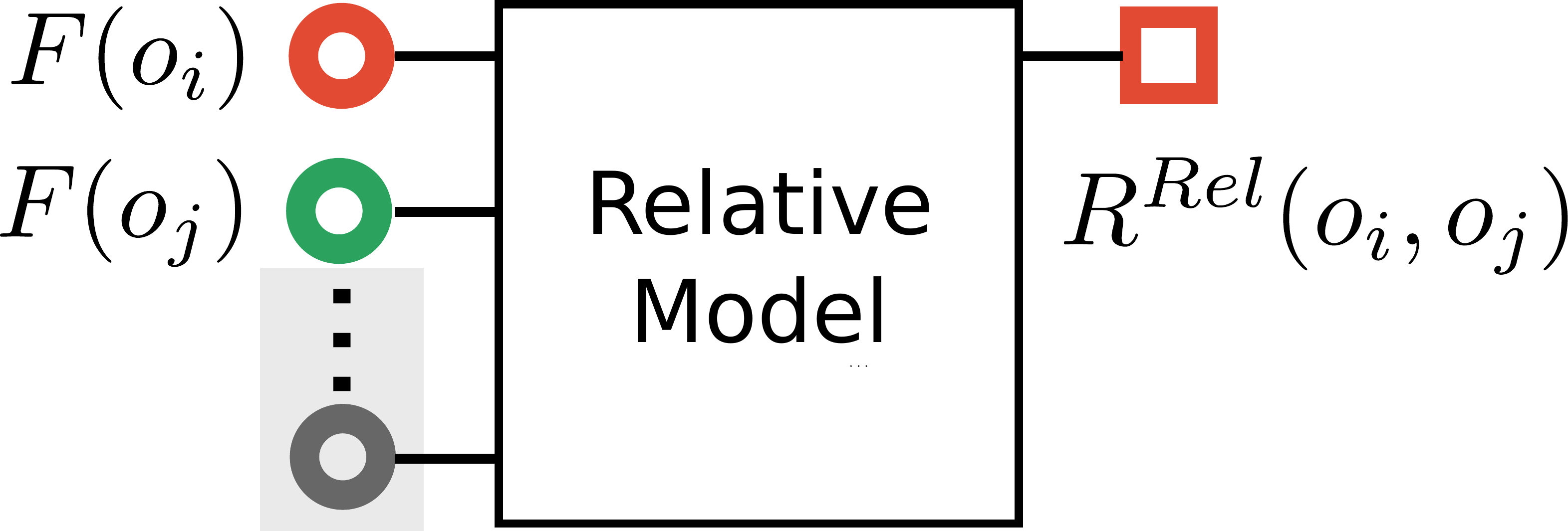}
	\end{subfigure}
    \caption{Models for spatial position and relation inference.}
    \label{fig:inference_models}
    \vspace{-0.5cm}
\end{figure}

Without any loss of generality, we impose conditional independence assumption to decompose the problem. We can solve equation~\eqref{eq:spatial_infer} by factorizing as 
\begin{equation}
\begin{aligned}
\label{eq:spatial_infer_decomp}
\underset{ \{ R^{Quad} , R^{Rel} \} }{\arg\max}  & \prod_{i=1}^{K} P \left( R^{Quad} (o_i) \mid F(o_1) , \cdots , F(o_K) \right) \\ & \prod_{i=1}^{K} \prod_{j=i+1}^{K} P \left( R^{Rel} (o_i, o_j) \mid F(o_1) , \cdots , F(o_K) \right)
\end{aligned}
\end{equation}
so that we can solve each factor independently. Using equation~\eqref{eq:spatial_infer_decomp}, we can have two inference problems for quadrants and relations. The conditional independence allows us to solve equation~\eqref{eq:spatial_infer_decomp} by:
\begin{equation}
\label{eq:multi_query}
\begin{aligned}
& \hat{R^{Quad} (o_i)} = \underset{ R^{Quad} }{\arg\max}~P \left( R^{Quad} (o_i) \mid F(o_1) , \cdots , F(o_K) \right) \\
& \hat{R^{Rel} (o_i, o_j)} \\ 
& = \underset{ R^{Rel} }{\arg\max}~P \left( R^{Rel} (o_i, o_j) \mid F(o_1) , \cdots , F(o_K) \right) 
\end{aligned}
\end{equation}
Figure~\ref{fig:inference_models} illustrates the two models that represent equation~\eqref{eq:spatial_infer_decomp}.
For an object $o_i$, we evaluate which quadrant it shall be in, $R^{Quad}(o_i)$, by the features of the object $F(o_i)$ and in the context of the features defined over all other objects $\{F(o_k) \mid k \in \{1, \cdots, K\} \setminus \{i\}\}$.
Similarly, we evaluate the spatial relation of $o_i$ to reference object $o_j$ where $i \neq j$ by features of both objects, $ F(o_i) $ and $ F(o_j) $, and in the context of the features defined over all other objects $\{F(o_k) \mid k \in \{1, \cdots, K\} \setminus \{i, j\}\}$.

We analyze our models' ability to represent a human's preference in this desk organization problem in the following aspects:
\begin{itemize}
	\item \emph{Accuracy:} How many spatial relations can the model correctly predict?
	\item \emph{Generalization:} Can this model generalize to different people's organizational preferences?
	\item \emph{Interpretability:} Can we determine from the model what rules a human uses when organizing objects?
	\item \emph{Satisfaction:} Is the person satisfied with the desk organization produced by the model?
\end{itemize}

In order to model the prediction defined in equation~\eqref{eq:multi_query}, we choose two canonical models, Random Forest~\cite{liaw2002classification} for classification precision and Markov Logic Network~\cite{richardson2006markov} for interpretability.

\subsection{Modeling using Markov Logic Networks}
\label{ssec:mln_approach}
The graph nature of MLN allows it to directly combine both quadrant and relative spatial relations in one model and allows for the querying of each type of relation independently. 
In order for the MLN to learn, it must first enumerate all predicate groundings using the specified domains. In order to limit the size of this space, we construct attribute domains \textsc{color} and \textsc{shape} to use a handful of unique values while \textsc{size}, \textsc{weight}, and \textsc{rigidity} use a binary classification.
In the MLN, we define the following domains:
\begin{center}
    \begin{tabular}{rcl}
        \textsc{quad} &=& \{1,2,3,4\}\\
        \textsc{dir} &=& \{\textsc{e, ne, n, nw, w, sw, s, se, in, none}\}\\
        \textsc{color} &=& \{\textsc{red, blue, black, green, yellow,}\\ &&\textsc{other}\}\\
        \textsc{shape} &=& \{\textsc{rectangle, cylinder, cube, other}\}\\
        \textsc{size} &=& \{\textsc{small, large}\}\\
        \textsc{weight} &=& \{\textsc{light, heavy}\}\\
        \textsc{rigidity} &=& \{\textsc{soft, hard}\}\\
        \textsc{utility} &=& \{1,2,3,4,5,6,7\}
    \end{tabular}
\end{center}

Initially, we provide the MLN with an unlearned set of formula expressed in first-order logic in terms of the predicates and domains above. We include formulae relating \textsc{color}, \textsc{shape}, \textsc{size}, \textsc{weight}, \textsc{rigidity}, and \textsc{utility} to \textsc{dir} and \textsc{quad}. During training, these formulae are expanded to include all possible groundings, after which, during the training process each formula receives a weight.

\subsection{Modeling using Random Forests}
Although less interpretable, random forests have more powerful representational abilities. As described earlier, we use two independent random forest models to capture all spatial relations. One model captured the quadrants for each object ($RF_{quad}$), and the other model captured the relative spatial relations between objects ($RF_{rel}$). The architecture of the 2 models is the same. The models use Gini impurity \cite{scikit-learn} as their criterion for splitting nodes. We keep the trees of the forests fully grown and unpruned, as our dataset was not so big as to put a cap on the memory consumption of the model. 
We define 20 estimators (20 decision trees comprising the forest) in each random forest.

We decided to split the spatial reasoning task into 2 models, which follow equation \eqref{eq:multi_query}; they deal with vectorized representations of features and spatial relations. Quadrant relations, and relative spatial relations operate in 2 separate domains, or classes, namely \textsc{quad} and \textsc{dir} as defined in \S~\ref{ssec:mln_approach}. So, following the domain design in MLN, we would need to enable multi-class classification. We describe in more detail how the two forests worked together to perform scene generation as well as how the accuracies of the two forests were computed.

For simplicity, the MLN predicate syntax as described in Section~\ref{ssec:mln_approach} was also used to describe scenes and objects with the Random Forest. For each object, the domain groundings (which act as features for the classifier) are converted into one-hot vectors. The vectors are concatenated to product input vectors for the random forest. In our usage, we restrict the set of desk scenes where $K = 7$. To generate scenes, $RF_{quad}$ is used to assign a quadrant to each of the 7 objects and then $RF_{rel}$ assigns cardinal relational directions between every pair of objects in the same quadrant.

\begin{figure*}[t!]
\centering
\begin{minipage}{0.48\linewidth}
\begin{subfigure}{\linewidth}
	\includegraphics[width=\linewidth]{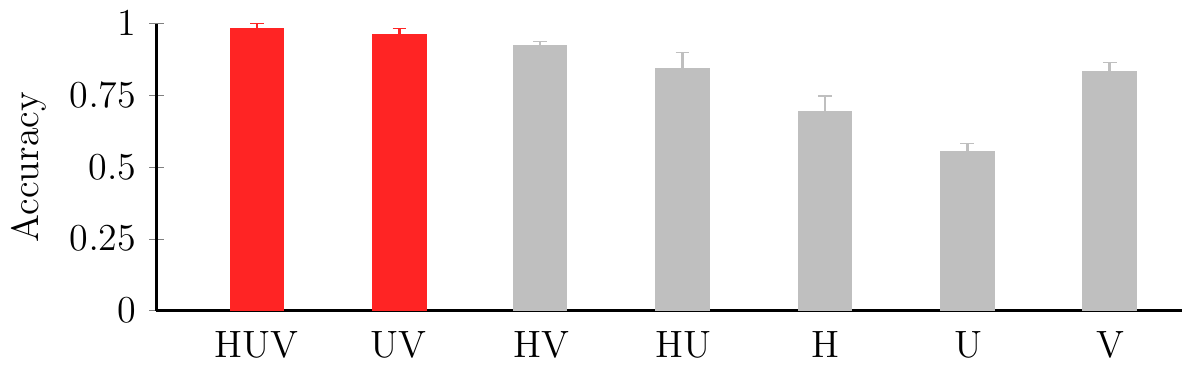}
	\caption{Random forests with different modalities}
	\label{plot:rf_modalities}
\end{subfigure}
\begin{subfigure}{\linewidth}
	\includegraphics[width=\linewidth]{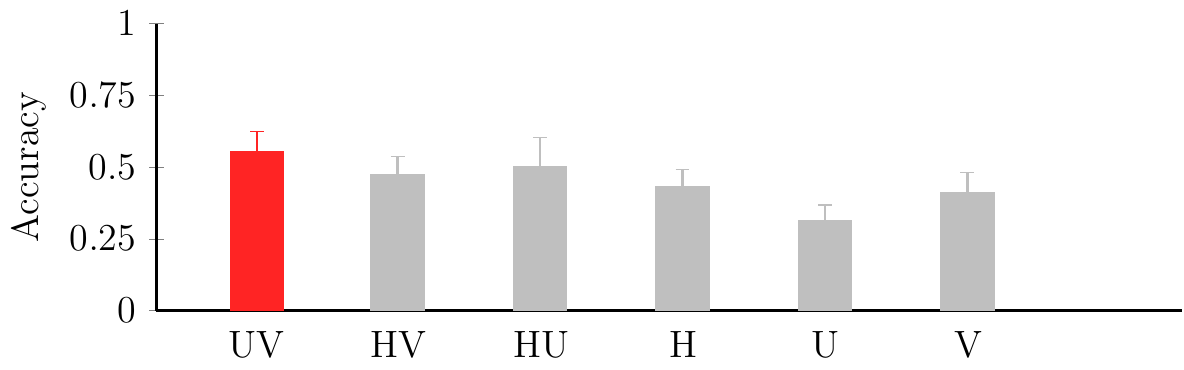}
	\caption{MLNs with different modalities}
	\label{plot:mln_modalities}
\end{subfigure}
\end{minipage}
\begin{minipage}{0.48\linewidth}
\begin{subfigure}{\linewidth}
	\includegraphics[width=\linewidth]{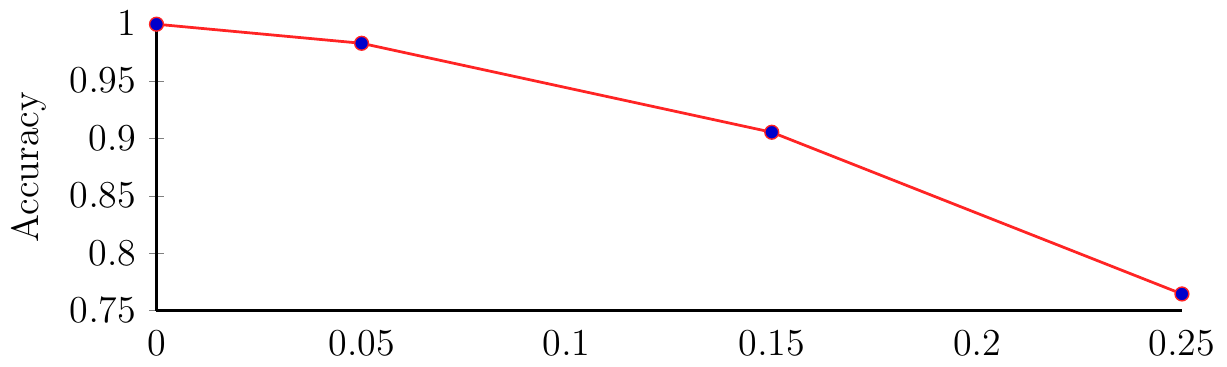}
	\caption{Increasing stochasticity}
	\label{plot:sim}
\end{subfigure}
\begin{subfigure}{\linewidth}
	\includegraphics[width=\linewidth]{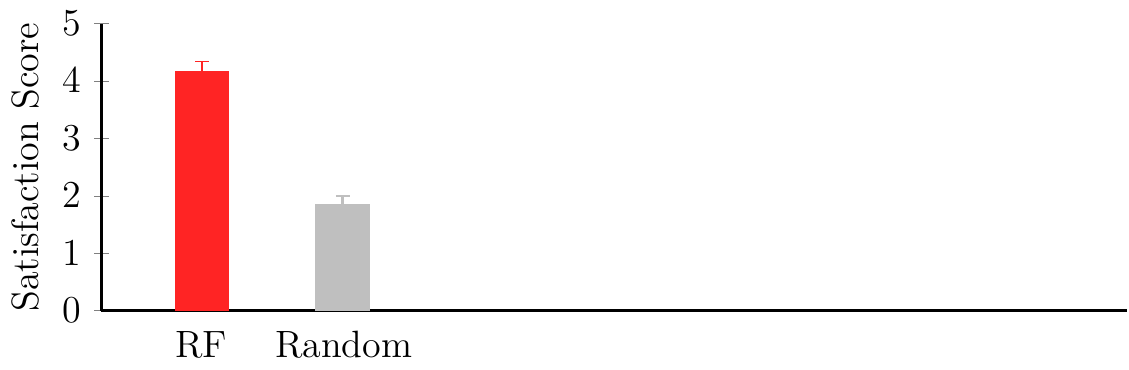}
	\caption{Participant satisfaction}
	\label{plot:follow_up}
\end{subfigure}
\end{minipage}
\caption{(a,b,c): Average fold accuracy during 5-fold cross-validation over 30 synthetic sample scenes. (d): Average participant satisfaction, rated on a 5-point Likert scale on follow up survey images organized by both trained random forest and a uniformly random quadrant model.}
\label{fig:sim_results}
\vspace{-0.5cm}
\end{figure*}

\section{Experiments and Results}
\label{sec:experiments}

\subsection{Experimental Setup}
For experimentation, we selected 17 objects that fit in the domains described in Section ~\ref{ssec:mln_approach}. These include typical desk objects such as a mouse, a box of paperclips, and a cellphone as well as more unusual objects such as an eraser cube, empty soda cans, and a Rubik's cube. These objects were chosen such that they shared properties in some modalities but differed in others. For example, the Rubik's cube and eraser cube both have the \textsc{shape} of \textsc{cube} and the \textsc{color} of \textsc{other}, but differ in their \textsc{rigidity}, where the Rubik's cube is \textsc{hard} and the eraser is \textsc{soft}. Some objects, such as the phone, only have one set of attributes while others, such as the dry erase marker, can have different attributes depending on which marker is used (red, blue, etc.). With our object set defined, we experimented with \textit{synthetic data}, \textit{human data} from multiple studies, and \textit{real-world robot demonstrations} in order to determine the representational power of our models.

\subsection{Synthetic data}
\label{ssec:simulation}

\subsubsection{Experimental Procedure}
We generated the synthetic data using a scene generator by picking objects at random with replacement from the set of all objects and positioning them on a desk programmatically according to a predefined list of constraints.

In collecting synthetic data, we generated 30 scenes. These scenes consisted of 6 to 9 objects chosen uniformly with replacement from the set of all objects. A quadrant annotator then produced a list of quadrant predicate groundings (both \textsc{quad} and \textsc{dir}), given the set of input objects, which described the position of every object. The groundings were chosen based on the \textsc{utility}, \textsc{color}, and \textsc{shape} of the provided objects.

\subsubsection{Results}
Using 5-fold cross-validation, MLNs were trained over the simulation set and used to predict object relations. We measured accuracy in terms of number of relational groundings (e.g., \textsc{dir}($o_0$, $o_5$, \textsc{N})) which were correctly predicted. Note that this relation means $o_0$ is to the North of $o_5$. Over the simulation set, MLN achieved 99\% accuracy, thus showing organizational spatial relations can be learned by one of our models.

To simulate uncertainty in annotation, ``stochasticity'' was added to the simulation. Several sets of true positional predicate groundings were generated with stochasticities $ p = 0.05 $, $ 0.15 $, and $ 0.25 $. It means that, with probability $p$ each object attribute was modified to a value other than its original (e.g., from \textsc{color}($o_i$, \textsc{red}) to \textsc{color}($o_i$, \textsc{blue})). The introduction of this parameter simulates a noisy or untrusted attribute annotator in order to determine ~\eqref{eq:spatial_infer}. Through 5-fold cross-validation with MLN, we found that the model was relatively robust in learning the simulated organizational strategy given small amounts of noise. As seen in Figure~\ref{plot:sim}, accuracy, hence learnability, decreases drastically with increasing stochasticity. %

\subsection{Human data: Initial study}
\label{ssec:human_data}

\subsubsection{Experimental Procedure}
For the initial study, we collected the human data during a human-participant study with 11 participants. The participants designed 30 scenes (similar to the experiments with synthetic data) by picking 7 objects with replacement. For each of the 30 scenes, we instructed each participant to organize the 7 scene objects on a desk divided into 4 quadrants only with the instruction being that objects may not span across quadrants and may not be on top of one another.

We also asked the study participants to subjectively characterize each unique object's \textsc{weight} and \textsc{size} in \{\textsc{light, heavy}\} and \{\textsc{small, large}\} categories respectively. These results were averaged over all participants to determine the canonical \textsc{weight} and \textsc{size} of each object. Each participant was also asked to rate the utility of each object on a 7-point Likert scale. These results were not averaged together; instead, when organizing a desk for a particular participant, each object was assigned the utility that the participant in question responded with. \textsc{rigidity} was determined by measuring the stiffness of each object with a spring scale and choosing a threshold such that half of the objects were \textsc{soft} and half were \textsc{hard}. Each object was measured on the surface where a human would normally grasp it. 

We trained two models over the same human dataset: 330 scenes, each with 7 objects, from 11 surveys with 5-fold cross-validation. In order to account for differences in an individual's organizational preferences, the dataset was partitioned by participant so models predicting for participant $n$ had only been trained on scenes from participant $n$. As described in the beginning of Section ~\ref{sec:experiments}, each participant's \textsc{utility} ratings for each object were used during training. In order to generate the true spatial relations for these scenes, during the study photos were taken of each scene from an overhead camera after organization. A positional annotator used hand-annotated masks for each object to determine which quadrant each was in, the pairwise cardinal relations for each object, and the \textsc{IN} relation if two masks sufficiently overlapped.

\begin{figure*}[t!]
\centering
\begin{tabular}{r|c|c|c|c|c}
     Truth &
     \includegraphics[width=.13\linewidth]{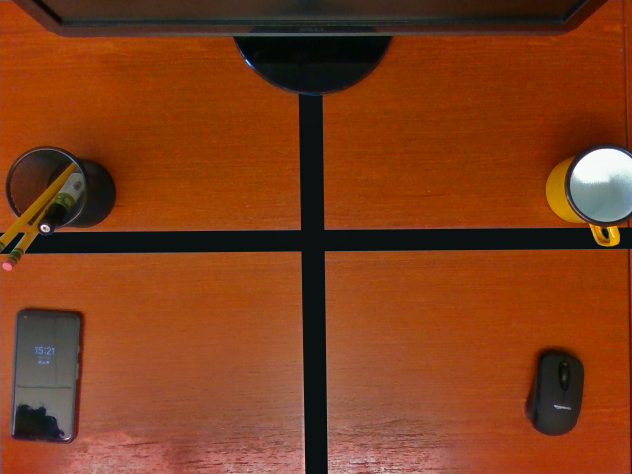} & \includegraphics[width=.13\linewidth]{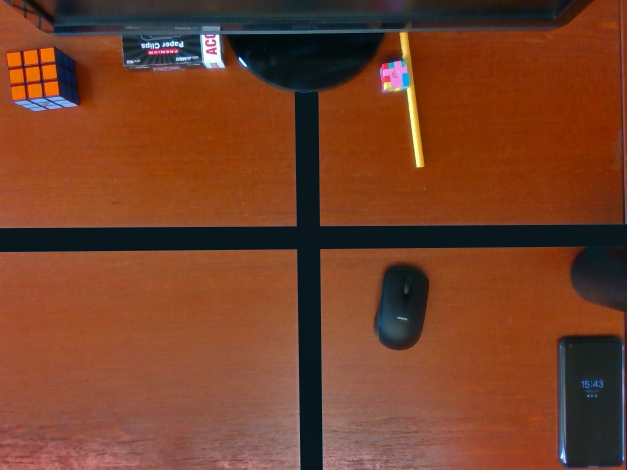} & \includegraphics[width=.13\linewidth]{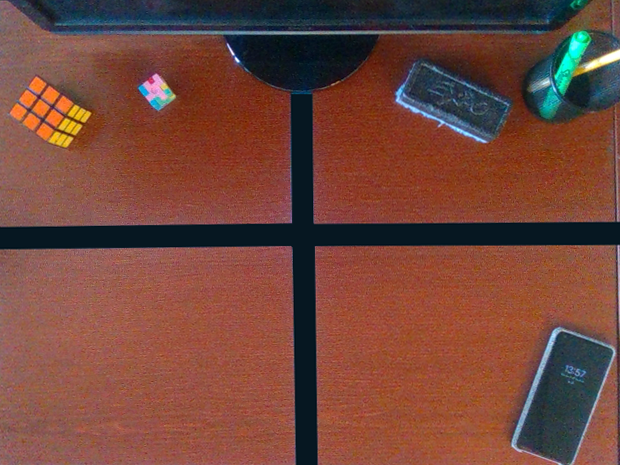} & \includegraphics[width=.13\linewidth]{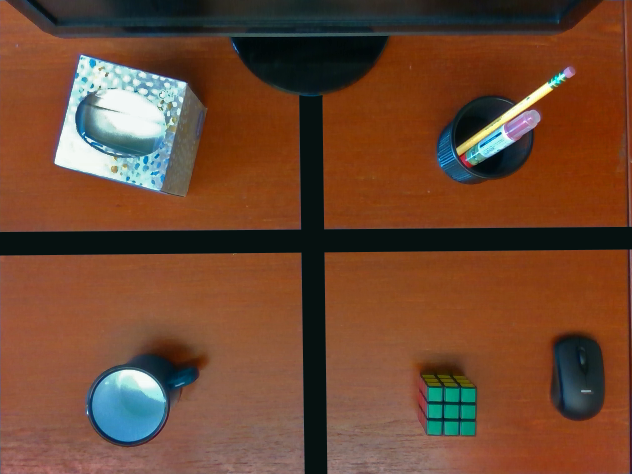} & \includegraphics[width=.13\linewidth]{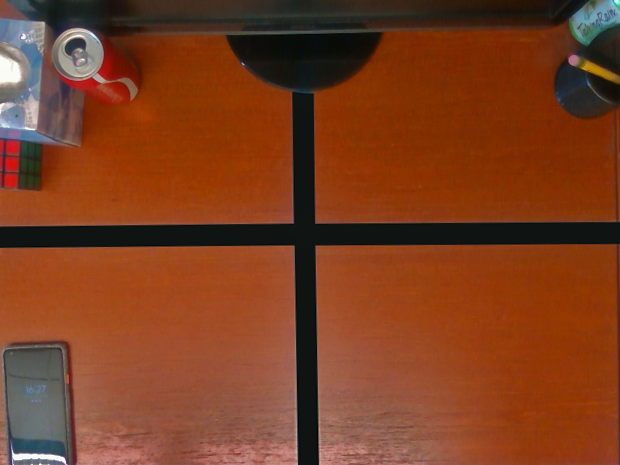}\\
     \hline
     RF &
     \includegraphics[width=.13\linewidth]{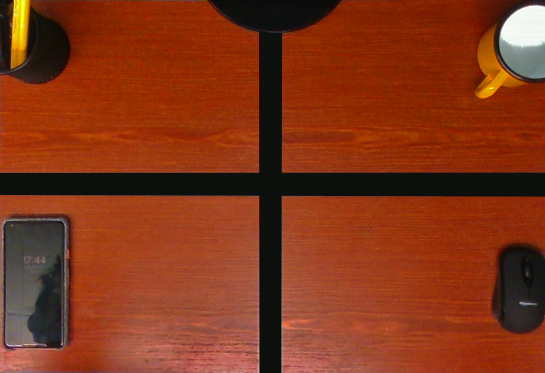} & \includegraphics[width=.13\linewidth]{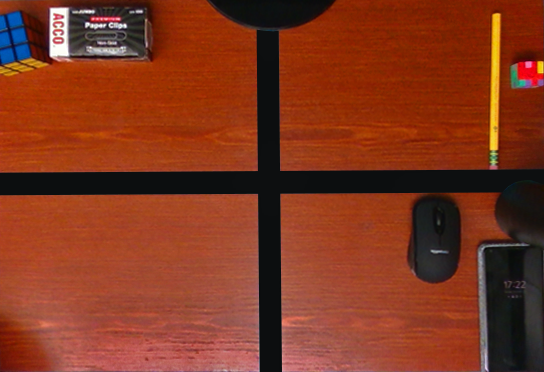} & \includegraphics[width=.13\linewidth]{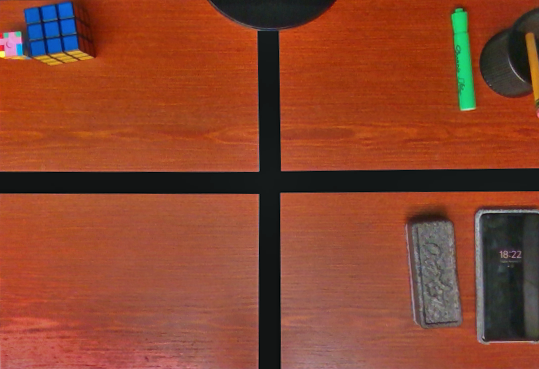} & \includegraphics[width=.13\linewidth]{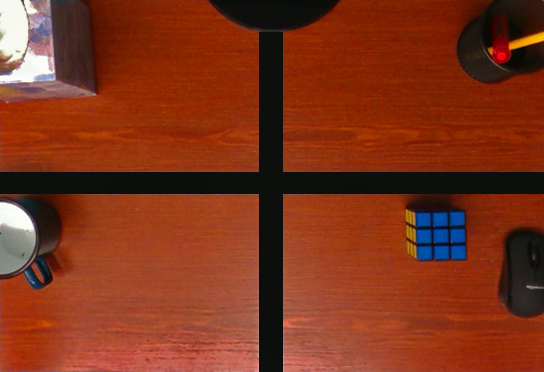} & \includegraphics[width=.13\linewidth]{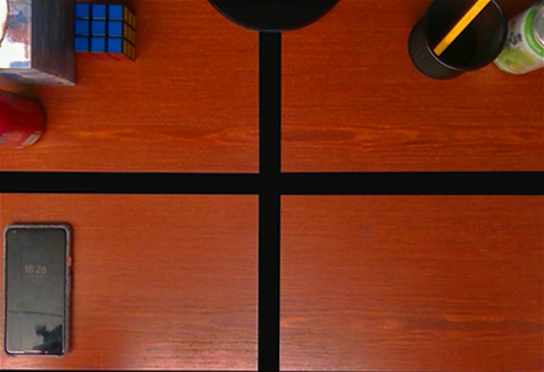}\\
     \hline
     Random &
     \includegraphics[width=.13\linewidth]{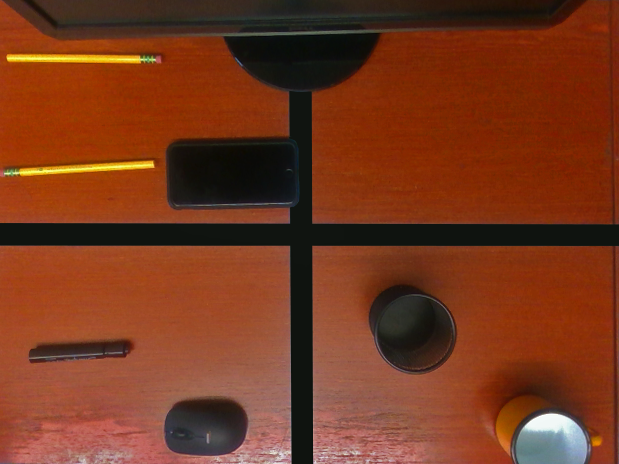} & \includegraphics[width=.13\linewidth]{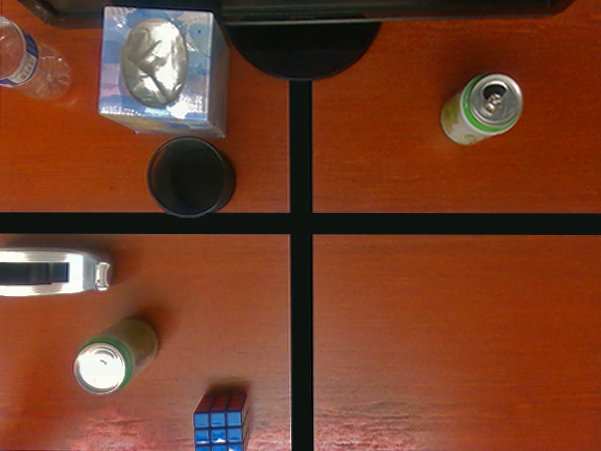} & \includegraphics[width=.13\linewidth]{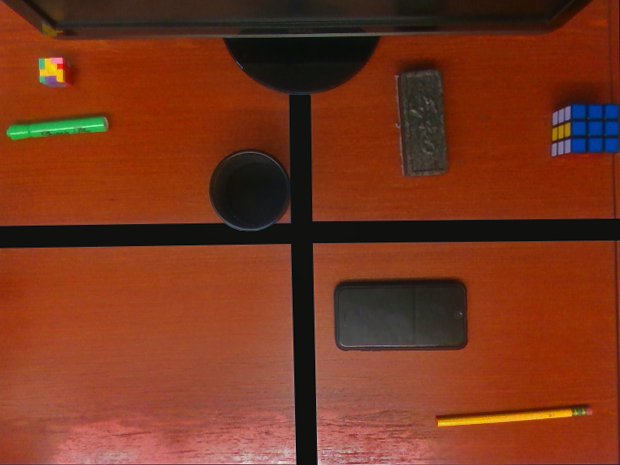} & \includegraphics[width=.13\linewidth]{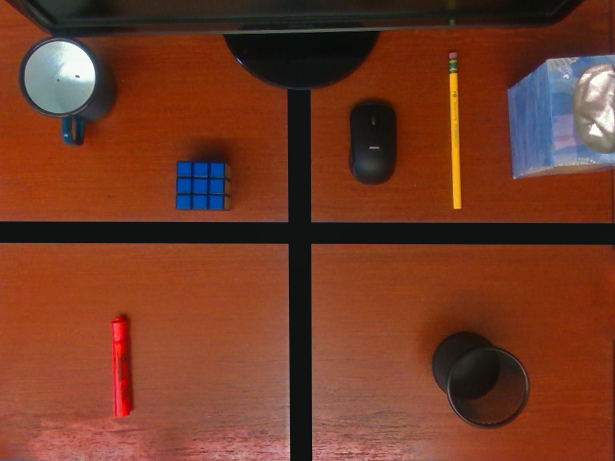} & \includegraphics[width=.13\linewidth]{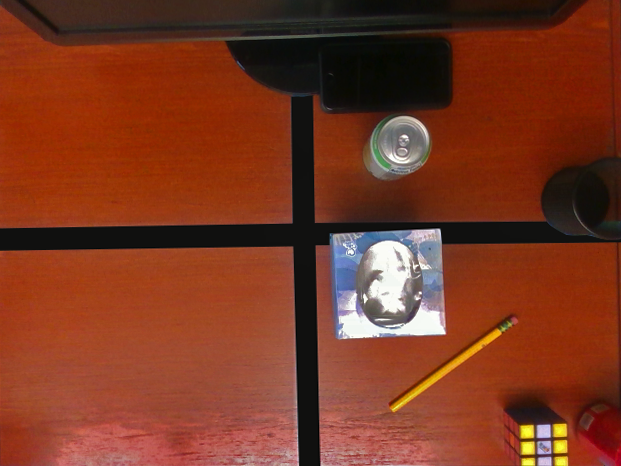}\\
     \hline
     & Scene 1 & Scene 16 & Scene 27 & Scene 28 & Scene 30
\end{tabular}
\caption{Five scenes from five different study participants used in the follow-up survey. Each scene was organized by the participant (Truth), a random forest model (RF), and a uniformly random positioner (Random).}
\label{fig:follow_up_org}
\end{figure*}

\subsubsection{Results}
Models were trained for each unique combination of the available modalities, resulting in 7 models: \textsc{HUV}, \textsc{HV}, \textsc{UV}, \textsc{HU}, \textsc{H}, \textsc{U}, \textsc{V}. Accuracy during cross-validation was measured for the MLN as described in Section~\ref{ssec:simulation}. For the random forest, the $RF_{quad}$ and $RF_{rel}$ cross-validation accuracies were averaged together with weights $K$ and ${K \choose 2}$ respectively where $K$ is the number of objects in the scene. This was done to make the results comparable with those of the MLN, as $K$ and ${K \choose 2}$ represent the ratio of \textsc{quad} and \textsc{dir} formula respectively.

As seen in Figure~\ref{plot:rf_modalities}, the Random Forest is able to achieve very high accuracy in cross-validation with nearly $95\%$ in average when using haptics, vision, and utility. With this decomposition, we can also observe that vision alone achieves $83\%$ accuracy in average, followed then by haptics and utility. Here we see the benefit of multimodal learning, as even with the introduction of one additional modality, haptics, to vision, we see the accuracy increase from $83\%$ to $92\%$ in average. The addition of our new modality, utility, increased accuracy further to $98\%$. Both of these are statistically significant increases, with p-values less than $0.05$, calculated via t-test. The t-statistic for the first is $16.945$ ($79.30$ degrees of freedom), and $27.399$ ($85.89$ degrees of freedom) for the second. When using both haptics and vision, the introduction of utility lead to a $6\%$ increase in accuracy. This, along with the $55\%$ accuracy when using utility alone, indicates that humans often take an object's usefulness into consideration when organizing items on a desk.

Although the MLN does surpass the RF in interpretability, it's representational abilities are far inferior. As explained in Section~\ref{ssec:mln_approach}, training the MLN requires expanding the provided formula with all combinations of domain groundings.
As a result, the number of \textsc{quad} formula is linear with domain size and number of modalities while the number of \textsc{dir} formula is multiplicative.
Due to this fact, the parameter space increases exponentially and convergence takes a long time (much more than the random forest). So, we omit this combination. Similarly, $10\%$ of non-\textsc{HUV} MLNs also failed to converge due to overflow and are therefore excluded from the data in Figure~\ref{plot:mln_modalities}. From this figure, we can again see that haptics and vision are among the most informative modalities, followed by utility.
However, the overall accuracy of the MLN model is much lower, with the highest accuracy being $71\%$ when using \textsc{HV} on participant 2's desk organizations.
Despite this, we again see that the interaction of multiple modalities yields higher overall predictive performance. Comparing utility with haptics plus utility, we see that the difference is statistically significant at $p < 0.05$ (t-statistic is $11.924$ with $84.842$ degrees of freedom). Note that these values are still much better than randomly guessing the positions of each of the objects. For example, in any one scene, there can be 7 unique objects. Each of these must be assigned 1 of 4 quadrants uniformly at random: the probability of being exactly correct is $1 / 4^7 = 0.00006$.

\subsubsection{Insights: Interpretability of MLNs}
Because MLN is programmed from first-order logics, the trained weight of the first-order formulae supports its interpretability. We can examine the weighted formula in the trained models to gain insights into one's organizational preferences. For example, with the highest weight of $16161$, the formula:

\begin{center}
     \textsc{utility}($o_1$, $5$) $\land$ \textsc{utility}($o_2$, $3$) $\land$ \textsc{dir}($o_1$, $o_2$, \textsc{NW})
\end{center}

informs us that utility is taken into account when positioning objects and that this participant prefers more useful objects to be positioned in front and to the right of less useful objects. For another participant, we see with weight $10130$, that:

\begin{center}
     \textsc{color}($o_1$, \textsc{blue}) $\land$ \textsc{color}($o_2$, \textsc{other}) $\land$ \textsc{dir}($o_1$, $o_2$, \textsc{SE})
\end{center}

This again gives us insight into the organizational preferences of this survey participant. Mainly that they tend to position \textsc{blue} objects behind and to the left of \textsc{other} objects.

\subsection{Human data: Follow-up Study}
\subsubsection{Experimental Procedure}
After the initial user study, which provided us with \emph{human data} to train the MLN and RF models on, we produced a numerical measure, namely accuracy, of the model's performance. However, human organizational habits are very subjective and multiple different arrangements of objects on a desk may be considered ``organized'' by different people or even the same person. To deal with this issue, we designed a follow-up survey to see how well our accuracy metrics matched ``human satisfaction''. This was largely motivated by our overarching goal: to build robots which can effectively operate in a three-dimensional world and interact with humans.

We sent follow-up surveys to each of the 11 participants of the original study.
For each survey, we randomly chose 5 of the 30 scenes from the original study. For each of the 5 scenes, we included in the survey 3 images of that scene:
\begin{enumerate}
    \item the scene as organized by the participant during the study.
    \item the scene as organized randomly.
    \item the scene as organized according to a Random Forest model's predictions.
\end{enumerate}

The original scenes are used as a reminder of what the study involved and how they originally organized each scene. They serves as a ``calibration'', so that their organizational schemes don't drift too far from the original study. For the random guesser, the 7 objects from the original scene were positioned uniformly randomly along the $x$ and $y$ axis of the desk plane. The randomly chosen location then inherently yielded the quadrant and relative spatial positions of each object. The purpose of these images is to provide a comparison to the human organizational scheme from before and the random forest organizational scheme. The scenes as organized by the random forest were included to measure our model's ability to learn a participant's organizational scheme. However, the random forest models (namely $RF_{quad}$ and $RF_{rel}$) used to generate this scene version were trained on scene data from \emph{all} participants. That is to say, the models for the follow-up survey were trained on $11 \times (30 - 5) = 275$ scenes, and then the scene generation was performed on 5 scenes per survey. 

In this capacity, the models are ``general'', as opposed to the ``personal'' models used in Section~\ref{sec:experiments}. Generalized models were chosen as opposed to tailored models to determine if, in addition to human satisfaction, there were cross-person patterns in how different people organized their desks and if the models could pick up on these patterns. This was directly motivated by real-life constraints of deploying agents with good enough \emph{priors} to perform personalized tasks, such as desk organization, out of the box before they have a chance to tailor their internal models to a particular person.

\subsubsection{Results: Generalization and Satisfaction}
In Figure~\ref{fig:follow_up_org}, we show $ 5 $ of the $ 30 $ scenes organized in three different ways for 5 of the participants. The random forest was able to successfully learn patterns in organizational choices that humans make. For example, in scene 28 it learned to place the mouse in the 4th quadrant and east of the Rubik's cube. It also successfully positioned the pencil and dry erase marker inside of the pen cup in scene 1.

Although the RF models used for generation of these organized scenes were generalized, and therefore different from the models that achieved high cross-validation accuracy in Section~\ref{ssec:human_data}, study participants rated the random forest organizations quite highly in terms of satisfaction. As seen in Figure~\ref{plot:follow_up}, participants were more satisfied with the scenes as organized by the random forest compared to those organized by the random guesser. The difference is statistically significant at $p < 0.05$ (t-statistic $ 68.784 $ with $105.726$ degrees of freedom).

\subsection{Robot Demonstration}
\begin{figure}[t!]
	\centering
	\includegraphics[width=0.45\linewidth]{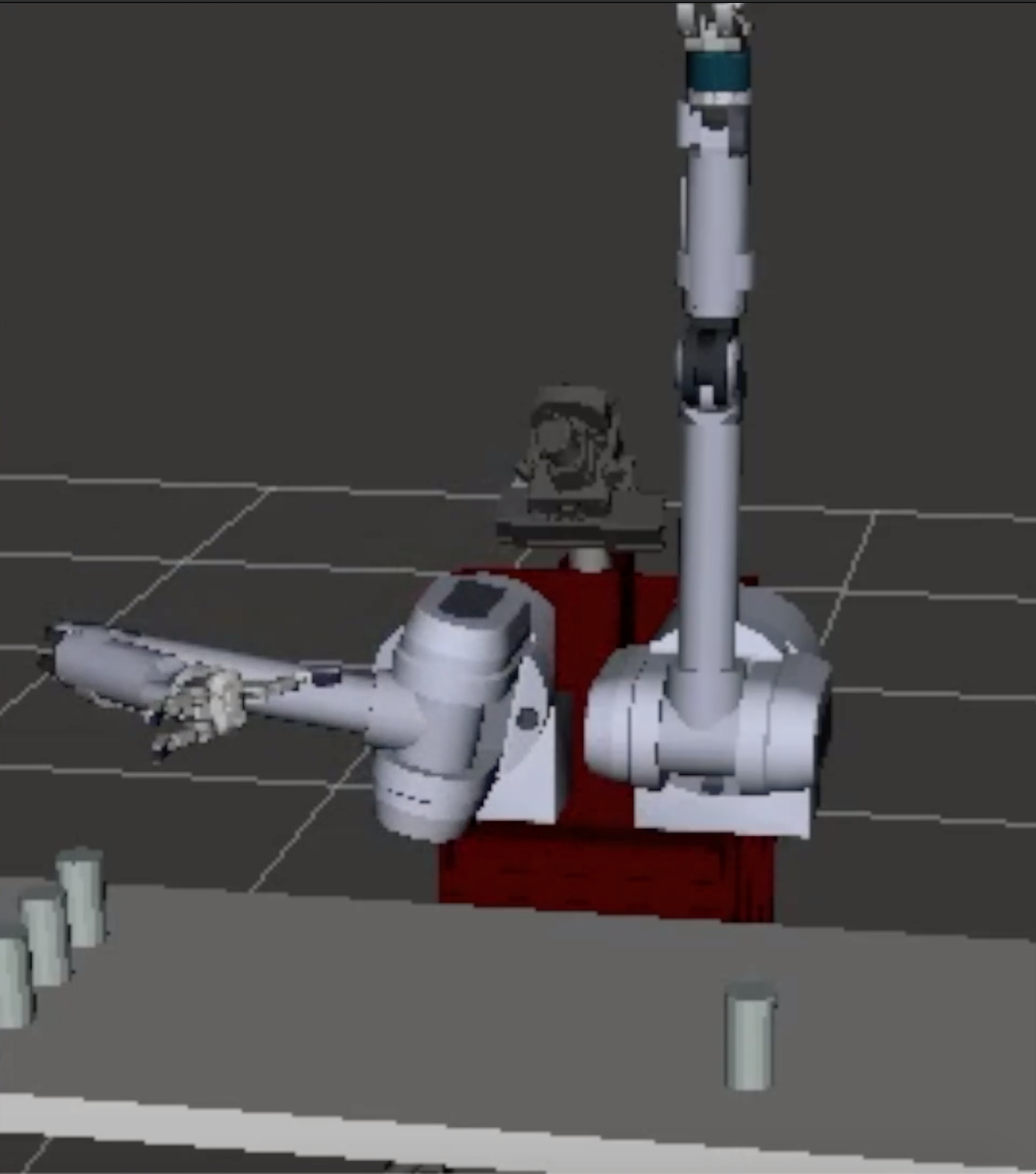}\,
	\includegraphics[width=0.45\linewidth]{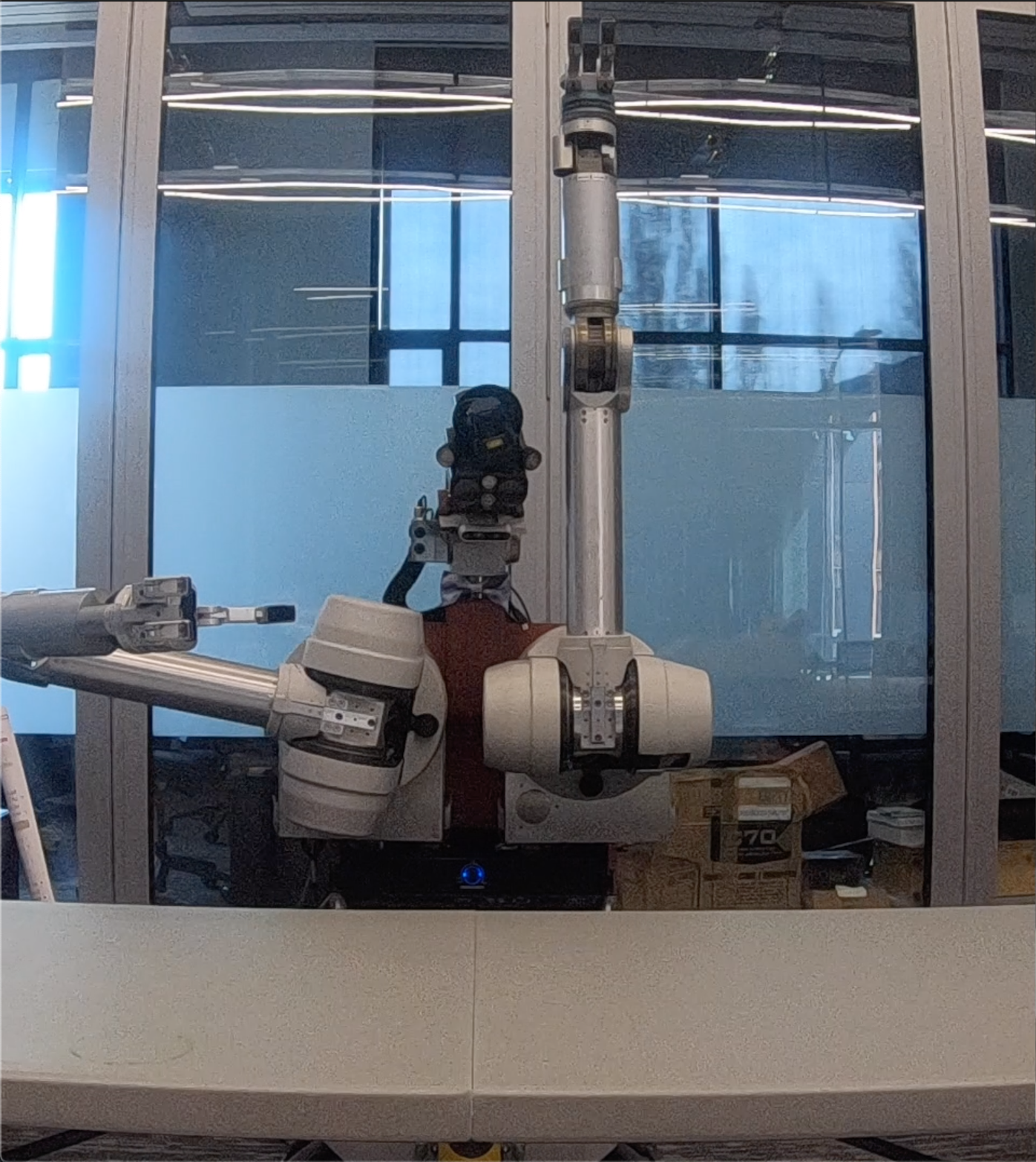}
	\caption{Robot trials in simulation and real-life. Video in \cite{Website}}
	\label{fig:robo}
	\vspace{-0.5cm}
\end{figure}

We used HERB 3.0 \cite{Srinivasa-2012-7533} to organize  $5$ soda cans in simulation (visualized using RViz), and in real. See Figure~\ref{fig:robo}. HERB is the Home Exploring Robot Butler; it is the robotic platform used for testing our models. HERB 3.0 has a mobile base and $2$ Barrett $7-$DOF WAM arms with Barrett hands; only arms and hands were used in manipulating objects in our experiment. It is also equipped with multiple laser rangefinders and cameras placed in various configurations (e.g. base, neck, etc.) to allow versatile perception capabilities; these were not used. Seeing as how our major goals were to analyze the spatial reasoning capabilities of the two models, and the advantage multimodal learning yields to it, We kept most of our work in ``simulation'' using cylindrical objects as a proof of concept. We also demonstrated the performance using an actual robot, HERB, performing a desk organization task, as a way of more tangibly demonstrating the above results. 

This trial demonstrates that with the proper modeling of spatial relations and using motion planning algorithms, robots performing such tasks as desk organization are feasible in home environments. As learned models mentioned above produce positions of objects according to spatial relations learned from human demonstrations, the focus of robot integration is moving objects from any initial start position to a goal position provided by our learned model. Thus, we designed our trial consisting of a start and goal position for each of the $5$ soda cans used in the experiments. The start positions were randomly selected on the right side of the table. The goal positions are extracted from running the random forest model. HERB was positioned near the table and used its right arm to successfully move all the soda cans to their goal positions \cite{Website}. 

\section{Discussion}
Random forests performed well but the weights of the model are difficult to interpret. 
Markov Logic Networks are mainly for performance and interpretability, as they perform well in simulation, and in the survey to some extent. It is easy to understand what an MLN learned as it simply consists of weighted first order logic formulae, which we designed to be very indicative of what features are being used to determine spatial relations.

The models do have their limitations as well. One limitation of MLNs is intractability.
If an MLN has too many formulae, inference becomes intractable, or takes prohibitively long. One limitation of the random forest model is that it requires fixed input dimensionality, meaning one trained on 8-object scenes cannot be used for inference on 9-object scenes. A realistic, deployable model would need to be able to handle scenes with any number of objects. Another limitation we faced with Random Forests was the presence of conflicts, that is, sometimes the random forests would suggest in one scene two conflicting relations. For example, say $RF_{quad}$ assigns \textsc{quad}($o_1$, $3$) and \textsc{quad}($o_2$, $3$) for two objects. $RF_{rel}$ could, in some instances, predict \textsc{dir}($o_1$, $o_2$, \textsc{N}) and \textsc{dir}($o_2$, $o_1$, \textsc{N}). This occurred because the $RF_{rel}$ does not take into account previous predictions in the scene; it predicts each relation independent of all other ones in the scene. However, we corrected for this by ignoring the conflicts in a manner consistent with how the program would have done it; once one relation is predicted, all other conflicting relations are null and void. Note that, in order to measure the capabilities of these models for capturing spatial relations, we ``turned off'' the noise, by using pristine, hand-labeled annotations from the surveys as opposed to running the MaskRCNN on them. In this, we make an assumption that the MaskRCNN can accurately detect each object and attribute annotators produce perfect annotations.

In the future, we plan to perform realistic robotic manipulation tasks in real homes. This would involve the creation of visual attribute annotators for color, shape, and size while haptic annotators would require use of a robotic arm and haptic sensor to pick up each object, as detected by MaskRCNN, to determine its rigidity and weight. This would involve some pipeline, similar to \ref{fig:pipeline}, which could also potentially involve natural language input to supplement or replace missing or inaccurate property annotations.

\bibliographystyle{IEEEtran}  %
\bibliography{mint-bibliography}  %

\begin{thebibliography}{10}
\providecommand{\url}[1]{#1}
\csname url@rmstyle\endcsname
\providecommand{\newblock}{\relax}
\providecommand{\bibinfo}[2]{#2}
\providecommand\BIBentrySTDinterwordspacing{\spaceskip=0pt\relax}
\providecommand\BIBentryALTinterwordstretchfactor{4}
\providecommand\BIBentryALTinterwordspacing{\spaceskip=\fontdimen2\font plus
\BIBentryALTinterwordstretchfactor\fontdimen3\font minus
  \fontdimen4\font\relax}
\providecommand\BIBforeignlanguage[2]{{%
\expandafter\ifx\csname l@#1\endcsname\relax
\typeout{** WARNING: IEEEtran.bst: No hyphenation pattern has been}%
\typeout{** loaded for the language `#1'. Using the pattern for}%
\typeout{** the default language instead.}%
\else
\language=\csname l@#1\endcsname
\fi
#2}}

\bibitem{fiorini2000cleaning}
P.~Fiorini and E.~Prassler, ``Cleaning and household robots: A technology
  survey,'' \emph{Autonomous robots}, vol.~9, no.~3, pp. 227--235, 2000.

\bibitem{dillmann2004teaching}
R.~Dillmann, ``Teaching and learning of robot tasks via observation of human
  performance,'' \emph{Robotics and Autonomous Systems}, vol.~47, no. 2-3, pp.
  109--116, 2004.

\bibitem{beetz2008assistive}
M.~Beetz, F.~Stulp, B.~Radig, J.~Bandouch, N.~Blodow, M.~Dolha, A.~Fedrizzi,
  D.~Jain, U.~Klank, I.~Kresse, \emph{et~al.}, ``The assistive kitchen—a
  demonstration scenario for cognitive technical systems,'' in \emph{Robot and
  Human Interactive Communication, 2008. RO-MAN 2008. The 17th IEEE
  International Symposium on}.\hskip 1em plus 0.5em minus 0.4em\relax IEEE,
  2008, pp. 1--8.

\bibitem{choi2009list}
Y.~S. Choi, T.~Deyle, T.~Chen, J.~D. Glass, and C.~C. Kemp, ``A list of
  household objects for robotic retrieval prioritized by people with als,'' in
  \emph{Rehabilitation Robotics, 2009. ICORR 2009. IEEE International
  Conference on}.\hskip 1em plus 0.5em minus 0.4em\relax IEEE, 2009, pp.
  510--517.

\bibitem{richardson2006markov}
M.~Richardson and P.~Domingos, ``Markov logic networks,'' \emph{Machine
  learning}, vol.~62, no. 1-2, pp. 107--136, 2006.

\bibitem{Breiman2001}
\BIBentryALTinterwordspacing
L.~Breiman, ``Random forests,'' \emph{Machine Learning}, vol.~45, no.~1, pp.
  5--32, Oct 2001. [Online]. Available:
  \url{https://doi.org/10.1023/A:1010933404324}
\BIBentrySTDinterwordspacing

\bibitem{he2017mask}
K.~He, G.~Gkioxari, P.~Doll{\'a}r, and R.~Girshick, ``Mask {R-CNN},'' in
  \emph{Computer Vision (ICCV), 2017 IEEE International Conference on}.\hskip
  1em plus 0.5em minus 0.4em\relax IEEE, 2017, pp. 2980--2988.

\bibitem{turaga2008machine}
P.~Turaga, R.~Chellappa, V.~S. Subrahmanian, and O.~Udrea, ``Machine
  recognition of human activities: A survey,'' \emph{IEEE Transactions on
  Circuits and Systems for Video technology}, vol.~18, no.~11, p. 1473, 2008.

\bibitem{chung2008daily}
P.-C. Chung and C.-D. Liu, ``A daily behavior enabled hidden markov model for
  human behavior understanding,'' \emph{Pattern Recognition}, vol.~41, no.~5,
  pp. 1572--1580, 2008.

\bibitem{Mansouri2013ARF}
M.~Mansouri and F.~Pecora, ``A representation for spatial reasoning in robotic
  planning,'' 2013.

\bibitem{spatialReasoningReview}
\BIBentryALTinterwordspacing
C.~Landsiedel, V.~Rieser, M.~Walter, and D.~Wollherr, ``A review of spatial
  reasoning and interaction for real-world robotics,'' \emph{Advanced
  Robotics}, vol.~31, no.~5, pp. 222--242, 2017. [Online]. Available:
  \url{https://doi.org/10.1080/01691864.2016.1277554}
\BIBentrySTDinterwordspacing

\bibitem{Kennedy2007SpatialRA}
W.~G. Kennedy, M.~D. Bugajska, M.~Marge, W.~Adams, B.~R. Fransen,
  D.~Perzanowski, A.~C. Schultz, and J.~G. Trafton, ``Spatial representation
  and reasoning for human-robot collaboration,'' in \emph{AAAI}, 2007.

\bibitem{paul2016efficient}
R.~Paul, J.~Arkin, N.~Roy, and T.~M~Howard, ``Efficient grounding of abstract
  spatial concepts for natural language interaction with robot manipulators,''
  2016.

\bibitem{webscalengrams}
S.~Li, G.~Kulkarni, T.~Berg, A.~Berg, and Y.~Choi,
  ``\BIBforeignlanguage{English (US)}{Composing simple image descriptions using
  web-scale n-grams},'' in \emph{\BIBforeignlanguage{English (US)}{CoNLL 2011 -
  Fifteenth Conference on Computational Natural Language Learning, Proceedings
  of the Conference}}, 2011, pp. 220--228.

\bibitem{yao2010i2t}
B.~Z. Yao, X.~Yang, L.~Lin, M.~W. Lee, and S.-C. Zhu, ``I2t: Image parsing to
  text description,'' \emph{Proceedings of the IEEE}, vol.~98, no.~8, pp.
  1485--1508, 2010.

\bibitem{bottomuptopdown}
\BIBentryALTinterwordspacing
P.~Anderson, X.~He, C.~Buehler, D.~Teney, M.~Johnson, S.~Gould, and L.~Zhang,
  ``Bottom-up and top-down attention for image captioning and {VQA},''
  \emph{CoRR}, vol. abs/1707.07998, 2017. [Online]. Available:
  \url{http://arxiv.org/abs/1707.07998}
\BIBentrySTDinterwordspacing

\bibitem{areasattentionic}
\BIBentryALTinterwordspacing
M.~Pedersoli, T.~Lucas, C.~Schmid, and J.~Verbeek, ``Areas of attention for
  image captioning,'' \emph{CoRR}, vol. abs/1612.01033, 2016. [Online].
  Available: \url{http://arxiv.org/abs/1612.01033}
\BIBentrySTDinterwordspacing

\bibitem{thomason2016learning}
\BIBentryALTinterwordspacing
J.~Thomason, J.~Sinapov, M.~Svetlik, P.~Stone, and R.~J. Mooney, ``Learning
  multi-modal grounded linguistic semantics by playing ``i spy'','' in
  \emph{Proceedings of the 25th International Joint Conference on Artificial
  Intelligence (IJCAI-16)}, New York City, 2016, pp. 3477--3483. [Online].
  Available:
  \url{http://www.cs.utexas.edu/users/ai-lab/pub-view.php?PubID=127564}
\BIBentrySTDinterwordspacing

\bibitem{taniguchi2016221}
\BIBentryALTinterwordspacing
A.~Taniguchi, T.~Taniguchi, and T.~Inamura, ``Simultaneous estimation of
  self-position and word from noisy utterances and sensory information,''
  \emph{IFAC-PapersOnLine}, vol.~49, no.~19, pp. 221 -- 226, 2016, 13th IFAC
  Symposium on Analysis, Design, and Evaluation ofHuman-Machine Systems HMS
  2016. [Online]. Available:
  \url{http://www.sciencedirect.com/science/article/pii/S2405896316321188}
\BIBentrySTDinterwordspacing

\bibitem{isobe2017learning}
S.~Isobe, A.~Taniguchi, Y.~Hagiwara, and T.~Taniguchi, ``Learning relationships
  between objects and places by multimodal spatial concept with bag of
  objects,'' in \emph{International Conference on Social Robotics}.\hskip 1em
  plus 0.5em minus 0.4em\relax Springer, 2017, pp. 115--125.

\bibitem{turk2014multimodal}
M.~Turk, ``Multimodal interaction: A review,'' \emph{Pattern Recognition
  Letters}, vol.~36, pp. 189--195, 2014.

\bibitem{ldamultimodal}
\BIBentryALTinterwordspacing
T.~Nakamura, T.~Araki, T.~Nagai, and N.~Iwahashi, ``Grounding of word meanings
  in latent dirichlet allocation-based multimodal concepts,'' \emph{Advanced
  Robotics}, vol.~25, no.~17, pp. 2189--2206, 2011. [Online]. Available:
  \url{https://doi.org/10.1163/016918611X595035}
\BIBentrySTDinterwordspacing

\bibitem{sisbot2007spatial}
E.~A. Sisbot, L.~F. Marin, and R.~Alami, ``Spatial reasoning for human robot
  interaction,'' in \emph{Intelligent Robots and Systems, 2007. IROS 2007.
  IEEE/RSJ International Conference on}.\hskip 1em plus 0.5em minus 0.4em\relax
  IEEE, 2007, pp. 2281--2287.

\bibitem{multimodalmachinelearning}
\BIBentryALTinterwordspacing
T.~Baltrusaitis, C.~Ahuja, and L.~Morency, ``Multimodal machine learning: {A}
  survey and taxonomy,'' vol. abs/1705.09406, 2017. [Online]. Available:
  \url{http://arxiv.org/abs/1705.09406}
\BIBentrySTDinterwordspacing

\bibitem{ban}
\BIBentryALTinterwordspacing
T.~Zhang, Y.-T. Li, and J.~P. Wachs, ``The effect of embodied interaction in
  visual-spatial navigation,'' \emph{ACM Trans. Interact. Intell. Syst.},
  vol.~7, no.~1, pp. 3:1--3:36, Dec. 2016. [Online]. Available:
  \url{http://doi.acm.org/10.1145/2953887}
\BIBentrySTDinterwordspacing

\bibitem{attentionSpatial}
\BIBentryALTinterwordspacing
D.~Kontogiorgos, ``Multimodal language grounding for improved human-robot
  collaboration: Exploring spatial semantic representations in the shared space
  of attention,'' in \emph{Proceedings of the 19th ACM International Conference
  on Multimodal Interaction}, ser. ICMI 2017.\hskip 1em plus 0.5em minus
  0.4em\relax New York, NY, USA: ACM, 2017, pp. 660--664. [Online]. Available:
  \url{http://doi.acm.org/10.1145/3136755.3137038}
\BIBentrySTDinterwordspacing

\bibitem{skovcaj2016integrated}
D.~Sko{\v{c}}aj, A.~Vre{\v{c}}ko, M.~Mahni{\v{c}}, M.~Jan{\'\i}{\v{c}}ek,
  G.-J.~M. Kruijff, M.~Hanheide, N.~Hawes, J.~L. Wyatt, T.~Keller, K.~Zhou,
  \emph{et~al.}, ``An integrated system for interactive continuous learning of
  categorical knowledge,'' \emph{Journal of Experimental \& Theoretical
  Artificial Intelligence}, vol.~28, no.~5, pp. 823--848, 2016.

\bibitem{nyga2014pr2}
D.~Nyga, F.~Balint-Benczedi, and M.~Beetz, ``Pr2 looking at things - ensemble
  learning for unstructured information processing with markov logic
  networks,'' in \emph{2014 IEEE International Conference on Robotics and
  Automation (ICRA)}, May 2014, pp. 3916--3923.

\bibitem{rf_alz}
K.~Gray, P.~Aljabar, R.~Heckemann, A.~Hammers, and D.~Rueckert, ``Random
  forest-based similarity measures for multi-modal classification of
  alzheimer's disease,'' \emph{NeuroImage}, vol.~65, 10 2012.

\bibitem{cvpr_rf}
H.~Kaya, F.~Gürpinar, and A.~A. Salah, ``Multi-modal score fusion and decision
  trees for explainable automatic job candidate screening from video cvs,'' pp.
  1651--1659, July 2017.

\bibitem{news_rf}
D.~Liparas, Y.~HaCohen-Kerner, A.~Moumtzidou, S.~Vrochidis, and
  I.~Kompatsiaris, ``News articles classification using random forests and
  weighted multimodal features,'' pp. 63--75, 2014.

\bibitem{malone1983people}
T.~W. Malone, ``How do people organize their desks?: Implications for the
  design of office information systems,'' \emph{ACM Transactions on Information
  Systems (TOIS)}, vol.~1, no.~1, pp. 99--112, 1983.

\bibitem{pantofaru2012exploring}
C.~Pantofaru, L.~Takayama, T.~Foote, and B.~Soto, ``Exploring the role of
  robots in home organization,'' in \emph{Proceedings of the seventh annual
  ACM/IEEE international conference on Human-Robot Interaction}.\hskip 1em plus
  0.5em minus 0.4em\relax ACM, 2012, pp. 327--334.

\bibitem{liaw2002classification}
A.~Liaw, M.~Wiener, \emph{et~al.}, ``Classification and regression by
  randomforest,'' \emph{R news}, vol.~2, no.~3, pp. 18--22, 2002.

\bibitem{scikit-learn}
F.~Pedregosa, G.~Varoquaux, A.~Gramfort, V.~Michel, B.~Thirion, O.~Grisel,
  M.~Blondel, P.~Prettenhofer, R.~Weiss, V.~Dubourg, J.~Vanderplas, A.~Passos,
  D.~Cournapeau, M.~Brucher, M.~Perrot, and E.~Duchesnay, ``Scikit-learn:
  Machine learning in {P}ython,'' \emph{Journal of Machine Learning Research},
  vol.~12, pp. 2825--2830, 2011.

\bibitem{Srinivasa-2012-7533}
S.~Srinivasa, D.~Berenson, M.~Cakmak, A.~C. Romea, M.~Dogar, A.~Dragan, R.~A.
  Knepper, T.~D. Niemueller, K.~Strabala, J.~M. Vandeweghe, and J.~Ziegler,
  ``Herb 2.0: Lessons learned from developing a mobile manipulator for the
  home,'' \emph{Proceedings of the IEEE}, vol. 100, no.~8, July 2012.

\bibitem{Website}
``Robot demonstration video,''
  \url{https://sites.google.com/cs.washington.edu/mintdeskorg/home},[Online;
  Retrieved on 10th May, 2019].

\end{thebibliography}

\end{document}